\documentclass[letterpaper]{article} 
\usepackage[draft]{aaai25}  
\usepackage{times}  
\usepackage{helvet}  
\usepackage{courier}  
\usepackage[hyphens]{url}  
\usepackage{graphicx} 
\usepackage{natbib}  
\usepackage{caption} 
\usepackage{algorithm}
\usepackage{algorithmic}

\usepackage{newfloat}
\usepackage{listings}

\usepackage{xcolor}
\usepackage{enumitem}
\usepackage{subcaption}
\usepackage{adjustbox}
\usepackage{multirow}
\usepackage{bbding}
\usepackage{epsfig}
\usepackage{amsmath}
\usepackage{amssymb}
\usepackage{makecell}
\usepackage{booktabs}
\newcommand{\blue}{\textcolor{blue}}
\DeclareCaptionStyle{ruled}{labelfont=normalfont,labelsep=colon,strut=off} 
\floatstyle{ruled}
\newfloat{listing}{tb}{lst}{}
\floatname{listing}{Listing}
\title{MM-Mixing: Multi-Modal Mixing Alignment for 3D Understanding}
\author{
    Jiaze Wang\equalcontrib\textsuperscript{\rm 1},
    Yi Wang\equalcontrib\textsuperscript{\rm 2},
    Ziyu Guo\textsuperscript{\rm 1},
    Renrui Zhang\textsuperscript{\rm 1},
    Donghao Zhou\textsuperscript{\rm 1}, \\
    Guangyong Chen\textsuperscript{\rm 3},
    Anfeng Liu\textsuperscript{\rm 2},
    Pheng-Ann Heng\textsuperscript{\rm 1}
}
\affiliations{

  \textsuperscript{\rm 1} The Chinese University of Hong Kong \quad
  \textsuperscript{\rm 2} Central South Unversity \quad 
  \textsuperscript{\rm 3} Zhejiang Lab
}

\begin{document}

\maketitle

\begin{abstract}
We introduce \textbf{MM-Mixing}, a multi-modal mixing alignment framework for 3D understanding. MM-Mixing applies mixing-based methods to multi-modal data, preserving and optimizing cross-modal connections while enhancing diversity and improving alignment across modalities.
Our proposed two-stage training pipeline combines feature-level and input-level mixing to optimize the 3D encoder. The first stage employs feature-level mixing with contrastive learning to align 3D features with their corresponding modalities. The second stage incorporates both feature-level and input-level mixing, introducing mixed point cloud inputs to further refine 3D feature representations. MM-Mixing enhances intermodality relationships, promotes generalization, and ensures feature consistency while providing diverse and realistic training samples.
We demonstrate that MM-Mixing significantly improves baseline performance across various learning scenarios, including zero-shot 3D classification, linear probing 3D classification, and cross-modal 3D shape retrieval. Notably, we improved the zero-shot classification accuracy on ScanObjectNN from 51.3\% to 61.9\%, and on Objaverse-LVIS from 46.8\% to 51.4\%. Our findings highlight the potential of multi-modal mixing-based alignment to significantly advance 3D object recognition and understanding while remaining straightforward to implement and integrate into existing frameworks.
\end{abstract}

\section{Introduction}
In the field of 3D vision, integrating multiple data modalities such as text, images, and point clouds has shown great potential for enhancing object recognition and scene understanding. This multi-modal approach is vital for applications in mixed reality ~\cite{dargan2023augmented,mendoza2023augmented}, autonomous navigation~\cite{chen20203d,tan2001exploring} and 3D scene understanding~\cite{armeni20163d,liu2021group,vu2022softgroup}, where accurate 3D perception is crucial. Recent advancements in multi-modal learning have underscored their capability in this domain, with notable contributions from seminal works like PointCLIP~\cite{zhang2022pointclip, zhu2023pointclip}, CLIP$^{2}$~\cite{zeng2023clip2}, ULIP~\cite{xue2023ulip,xue2023ulip2}, OpenShape~\cite{liu2024openshape}, and TAMM~\cite{zhang2024tamm}. These studies have demonstrated the effectiveness of leveraging text, images, and point clouds to improve 3D object recognition and understanding.

\begin{figure}[t]
    \centering
        \includegraphics[width=\linewidth]{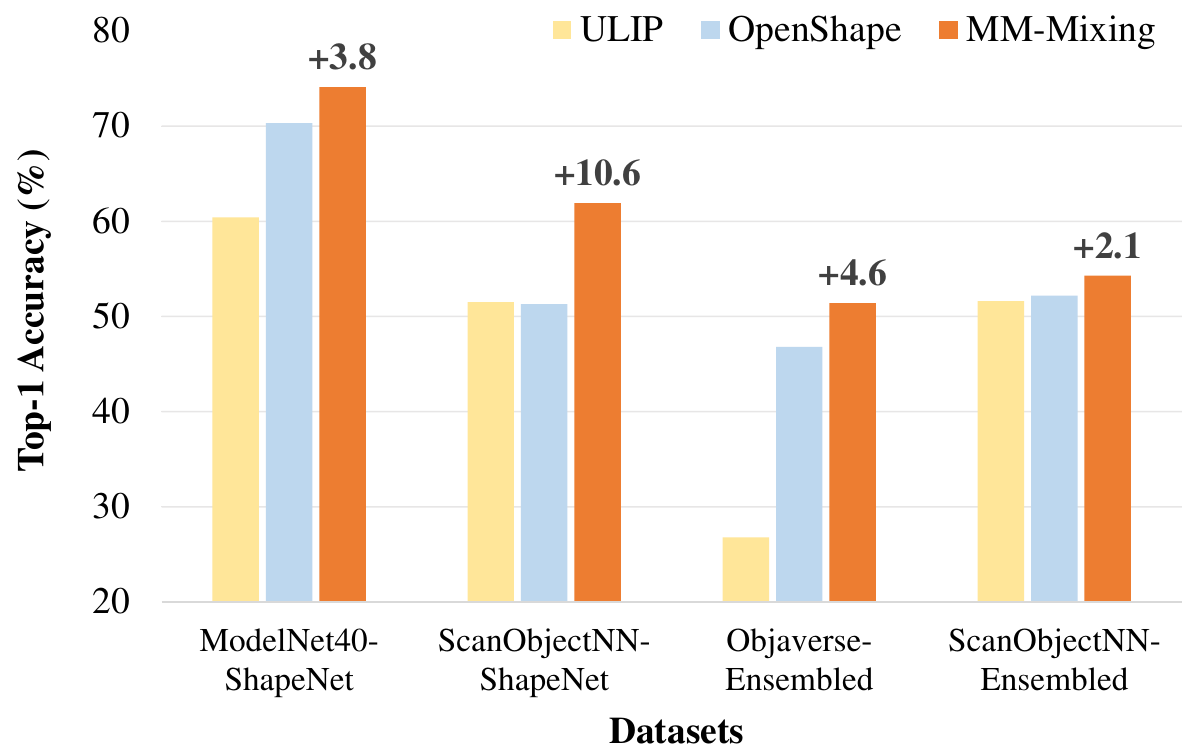}
    \caption{\textbf{Performance comparison with previous methods}. MM-Mixing achieves better performance than previous pre-training methods across various datasets with the same backbone Point-BERT. ``ModelNet40-ShapeNet" represents the model is pretrained on ShapeNet and evaluated on ModelNet40, similarly for other dataset combinations.}
    \label{fig: performance}
    \vspace{-12pt}
\end{figure}

However, a significant challenge remains in effectively aligning and utilizing these heterogeneous data sources to optimize model performance. 
With recent advancements in 3D vision, there's a growing emphasis on multi-modal learning approaches. These frameworks are becoming increasingly crucial, especially when it comes to processing and learning from multi-modal data, which integrates textual information, 2D images, and 3D point cloud data.
Despite the success of these approaches, there is a notable gap in the literature regarding multi-modal data augmentation. The cohesive augmentation of triplets has the potential to unlock further performance improvements by enriching the diversity of data and promoting better alignment across modalities. This presents a promising avenue for research to explore comprehensively the benefits of multi-modal learning frameworks.

In previous studies, many mixing-based data augmentation
methods have been proposed for point cloud~\cite{kim2021point,rao2021randomrooms,lee2022sagemix}. Mixing-based methods like PointCutMix~\cite{zhang2022pointcutmix} and PointMixup~\cite{chen2020pointmixup} enhance training data diversity through techniques such as region splicing and feature interpolation. By introducing controlled perturbations and heterogeneity into the training process, these approaches enable models to learn invariant and discriminative features, thereby improving their robustness and generalization to diverse and unseen data distributions~\cite{umam2022point,kim2021point,wang2024pointpatchmix}. 

However, the potential of mixing-based methods in multi-modal scenarios remains largely unexplored. Integrating mixing-based techniques with multi-modal alignment could enhance multi-modal learning by generating diverse feature spaces, fostering robust cross-modal correspondences, and revealing invariant features across modalities. This leads to an important question: \emph{Can we design a simple yet effective framework that improves alignment quality and stability while enhancing model generalization through augmented, coherent multi-modal representations?}
 
To address this issue, we introduce \textbf{MM-Mixing}, a multi-modal approach for 3D understanding that integrates mixing-based methods with multi-modal triplet data. Our two-stage training pipeline combines feature-level and input-level mixing to optimize the 3D encoder, enhancing intermodality relationships and promoting generalization.
In the first stage, MM-Mixing leverages feature-level mixing and contrastive learning to align mixed 3D features with their corresponding modalities. This mixing-based alignment strategy fosters consistency across different modalities and significantly enhances the 3D encoder's cross-modal understanding. Specifically, by aligning point cloud mixed features with text mixed features, we capture semantic information that provides a contextual understanding of the 3D shapes. Additionally, aligning point cloud mixed features with image mixed features bolsters the capture of intricate visual details and spatial relationships. This dual alignment of mixed features not only ensures cross-modal consistency but also amplifies the 3D encoder's ability to understand and represent complex, multi-modal data effectively.
The second stage incorporates feature-level and input-level mixing, introducing mixed point cloud inputs to refine 3D feature representations further. By aligning mixed point cloud features with feature-level mixed point cloud features, we enhance the network's ability to capture and represent variations and nuances within the data, resulting in more robust and discriminative feature representations. This stage generates diverse and realistic samples that enhance the 3D encoder's ability to generalize across different datasets.

By seamlessly integrating these methods, MM-Mixing significantly boosts the baseline model's performance across various settings, including zero-shot 3D classification, linear probing 3D classification, and cross-modal 3D shape retrieval, while remaining straightforward to implement and integrate into existing 3D understanding frameworks.
Our main contributions can be summarized as follows:

\begin{itemize}[noitemsep,topsep=0pt,leftmargin=*]
\item We introduce MM-Mixing, a novel multi-modal mixing alignment framework specifically designed for multi-modal data, addressing a previously unexplored issue in 3D understanding, which can be easily integrated with existing frameworks.
\item An efficient two-stage framework is proposed that integrates feature-level and input-level augmentation to optimize the 3D encoder, enhance cross-modal relationships, and promote generalization.
\item Our MM-Mixing not only strengthens the 3D understanding of models but also significantly enhances cross-dataset generalization, demonstrating exceptional performance in downstream tasks such as zero-shot 3D classification, linear probing 3D classification, and cross-modal retrieval.
\end{itemize}

\section{Related Works}
\noindent\textbf{3D Understanding.}
Understanding 3D structures is a crucial aspect of computer vision~\citep{peng2023openscene,qi2023contrast,rozenberszki2022language,zhang2022glipv2,zhang2023clip}.
Recent developments in 3D understanding have largely focused on leveraging advanced representation learning techniques~\citep{abdelreheem2023satr,achituve2021self,achlioptas2018learning,aneja2023clipface,deng2018ppf,hess2023masked,guo2023point}. Three primary methodologies have emerged: projecting-based methods where 3D point clouds are projected into various image planes~\citep{su2015multi,kanezaki2018rotationnet,goyal2021revisiting,chen2017multi}, voxel-based methods which transform the point clouds with 3D voxelization~\citep{song2017semantic,riegler2017octnet,canfes2023text}, and direct modeling of 3D point clouds with point-centric architectures~\citep{qian2022pointnext, ma2022rethinking}. These approaches highlight the use of specialized models like SparseConv~\citep{choy20194d} for efficiently handling sparse voxel data, and Transformer-based models~\citep{guo2023joint,zhang2023learning} such as Point-MAE~\citep{pang2022masked}, Point-M2AE~\citep{zhang2022point} and Point-BERT~\citep{yu2022point} for leveraging self-supervised learning paradigms.
Moreover, the integration of image-language models like CLIP~\citep{radford2021learning} into 3D shape understanding represents a significant trend~\citep{zhang2022pointclip,zhu2023pointclip,zeng2023clip2,huang2023clip2point,liu2024openshape,zhang2024tamm,chen2023clip2scene,wang2021category,pnn,zhu2024no}. Models are trained to align 3D shape embeddings with CLIP’s language and/or image embeddings through multimodal contrastive learning~\citep{yuan2021multimodal,ding2023pla,ha2022semantic,hegde2023clip,hong2022avatarclip,huang2024joint, jatavallabhula2023conceptfusion,chen2024signvtcl,liang2022mind,liu2023partslip,zhang2023learning}. This allows for zero-shot 3D classification and improves the robustness of shape representations. Notably, advancements such as ULIP~\citep{xue2023ulip,xue2023ulip2}, I2P-MAE~\cite{zhang2023learning}, and OpenShape~\citep{liu2024openshape} have sought to refine this approach by optimizing the distillation of CLIP features into 3D representations and expanding training datasets for more generalizable learning outcomes.

\noindent\textbf{3D Mixing-based Augmentation.}
In the realm of 3D mixing-based methods, significant strides have been made to enhance the diversity and quality of augmented point cloud data. Traditional techniques primarily involved simple transformations such as rotation, scaling, and jittering at the point level~\citep{ren2022benchmarking,qi2017pointnet,qi2017pointnet++,goyal2021revisiting}. However, recent innovations have introduced more sophisticated methods that preserve or even enhance the structural integrity of point clouds while introducing variability. For instance, PointAugment~\citep{li2020pointaugment} optimizes both enhancer and classifier networks to generate complex samples, while techniques like Mixing-based augmentation~\citep{chen2020pointmixup,zhang2022pointcutmix,wang2024pointpatchmix,lee2021regularization} employ strategies from the 2D domain, such as optimal linear interpolation and rigid transformations, to mix multiple samples effectively.
Furthermore, the advent of Transformer-based methods and attention mechanisms in point cloud processing has opened new possibilities for data augmentation. PointWOLF~\citep{kim2021point} introduces multiple weighted local transformations, and PointMixSwap~\citep{umam2022point} utilizes an attention-based method to swap divisions across point clouds, adding a layer of complexity and diversity. Additionally, with the development of PointPatchMix~\citep{wang2024pointpatchmix}, point cloud mixing occurs at the patch level, which can generate more realistic  data with the self-attention mechanism. 

\begin{figure*}[t]
    \centering
        \includegraphics[width=\linewidth]{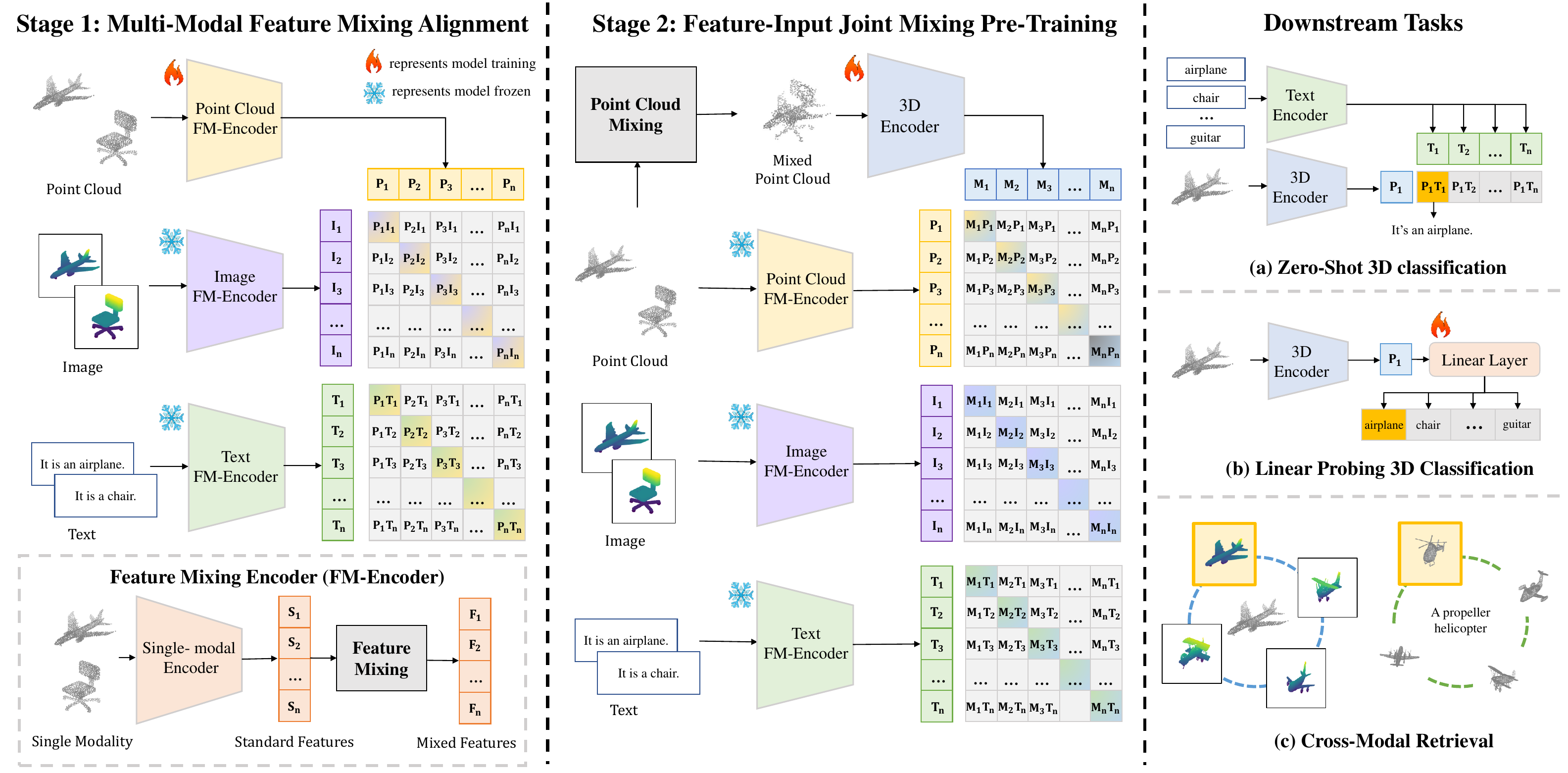}
\caption{\textbf{The overall scheme of MM-Mixing.} MM-Mixing consists of two stages. In the first stage, the point cloud FM-Encoder is trainable, while the image and text FM-Encoders are pre-trained and frozen. Feature embeddings are extracted for contrastive learning with the 3D features.
In the second stage, we initialize a new trainable 3D encoder. All FM-Encoders remain frozen. Two input point clouds are mixed using FPS and point-level mixing, and then fed into the 3D encoder. Then we adopt contrastive learning to align the features of mixed point clouds with mixed feature representations of all three modalities.}
    \label{fig: main}
    \vspace{-6pt}
\end{figure*}

\section{Method}

The overall MM-Mixing pipeline is shown in Figure \ref{fig: main}. We first review the problem definition to establish the context of our approach. Then, we introduce our mixing-based alignment strategy specifically designed for point clouds, images, and texts, which enhances the variability and robustness of the training data. Finally, we detail the MM-Mixing framework, demonstrating how our method integrates seamlessly into existing frameworks.

\subsection{Problem Definition}
Given a set of \( K \) triplets \(\{(P_i, I_i, T_i)\}_{i=1}^K\), where \( P_i \) is a 3D point cloud, \( I_i \) represents the corresponding image produced by projecting the 3D point cloud \( P_i \) into 2D from an arbitrary perspective, and \( T_i \) denotes the associated text generated using advanced vision-language models such as BLIP~\citep{li2022blip}, the objective is to learn high-quality 3D representations from these triplets.
Following ULIP~\citep{xue2023ulip} and OpenShape~\citep{liu2024openshape} which leverage the CLIP~\citep{radford2021learning} model, we enhance this framework by incorporating mixing-based methods. Specifically, the 3D features of the mixed point cloud $m_i^{M}=E_P(I_M(P_i, P_j))$ are obtained by passing two point clouds sequentially through the input-level mixing $I_M$ and the 3D encoder $E_P$. The corresponding mixed features of the point cloud modality $m_i^{P}=F_M(E_P(P_i),E_P(P_j))$, the mixed features of the image modality $m_i^{I}=F_M(E_I(I_i),E_I(I_j))$, and the mixed features of the text modality $m_i^{T}=F_M(E_T(T_i),E_T(T_j))$ are generated by passing the features produced by the trained modality-specific encoders $E_P$, $E_I$ and $E_T$ through the feature-level mixing $F_M$, respectively. During the optimization of the 3D encoder $E_P$, contrastive learning is used to align the 3D features of the mixed point cloud $m_i^{M}$ with the mixed features of the three modalities $m_i^{P}$, $m_i^{I}$, $m_i^{T}$.

\subsection{Multi-Modal Mixing}
We adopt two kinds of mixing methods for multi-model data, including feature-level mixing and input-level mixing.

\noindent\textbf{Feature-level mixing.}
Feature-level mixing augments the features by combining features from two different inputs. This process involves first passing each input through the network independently to extract their respective features. Specifically, the first input is fed into the network, which processes it and extracts its feature vector \( f_i \). Similarly, the second input is also passed through the network, resulting in the extraction of its feature vector \( f_j \). Then the features are combined using a mixing operation to create a new, combined feature vector $m_i$, which can be expressed as:
\begin{small}
\begin{equation}
m_i = \lambda f_i + (1 - \lambda) f_j.
\end{equation}
\end{small}

\noindent\textbf{Input-level mixing.}
For input-level mixing, we follow PointCutMix~\citep{zhang2022pointcutmix}, which generates a new training point cloud $\tilde{p}$ from a pair of point clouds $p_{1}$ and $p_{2}$.
The combination process of input-level augmentation is defined as follows:
\begin{small}
\begin{align}
M & = S \odot P_{1} + (1 - S) \odot P_{2}, \\ 
\lambda & = \sum S/N,
\end{align}
\end{small}where $M$ is the mixed point cloud, $S \in \{0, 1\}^{N}$ indicates which sample each point belongs to, $\odot$ represents element-wise multiplication, and $\lambda$ is sampled from a beta distribution $Beta(\beta, \beta)$. This implies that $\lfloor \lambda N \rfloor$ points are selected from $p_{1}$, and $N - \lfloor \lambda N \rfloor$ points are selected from $p_{2}$.

Feature-level mixing operates on the encoded feature vectors, inducing implicit changes in the high-dimensional space. This allows for efficient data augmentation under cross-modal conditions, ensuring consistency of the augmented features across different modalities. In contrast, input-level augmentation directly manipulates the raw data, generating concrete and intuitive mixed samples. These realistic samples, which are both challenging and diverse, help the model better understand 3D shapes in downstream tasks. MM-Mixing combines these two augmentation strategies, achieving dual enhancement between raw data and latent features, thereby significantly improving the model's generalization ability.

\subsection{MM-Mixing Framework}
MM-Mixing refines feature representations through a combination of contrastive learning and mixing-based augmentation techniques, which improves the encoder's ability to generalize and discriminate between different classes through a two-stage training framework.

As shown in Figure \ref{fig: main}, in the first stage, the point cloud Feature Mixing Encoder (FM-Encoder) is trainable, we freeze the image and text Feature Mixing Encoders (FM-Encoders), which are a combination of a single-modal encoder from CLIP~\citep{radford2021learning} with a feature mixing module. Initially, point clouds are fed into the trainable point cloud Feature Mixing Encoder (FM-Encoder) to obtain 3D mixed feature embeddings. Concurrently, corresponding images and textual descriptions are processed through the frozen image and text Feature Mixing Encoders (FM-Encoders) to extract image and text mixed feature embeddings. These extracted 3D, image, and text features are then combined to mixed feature triplets. Employing a contrastive learning objective, the mixed 3D features are aligned with the image and text mixed features. This encourages the point cloud Feature Mixing Encoder (FM-Encoder) to learn a feature space that is consistent with the representations of the frozen encoders from other modalities, enhancing its ability to discriminate between different 3D objects. The Stage 1 corresponding contrastive loss $L^{S1}$ is calculated as:
\begin{small}
\begin{equation}
F(x, y) = \log\frac{\exp(x \cdot y / \tau)}{\sum_{j} \exp(x_j \cdot y_j / \tau)},
\end{equation}
\end{small}

\begin{small}
\begin{equation}
\begin{split}
L^{S1} = -\frac{1}{4n} \sum_{i} (
& F(m_i^{P}, m_i^{I})
+ F(m_i^{I}, m_i^{P}) + \\ 
&F(m_i^{P}, m_i^{T})
+ F(m_i^{T}, m_i^{P})
),
\end{split}
\end{equation}
\end{small}where $n$ is the number of mixed features in a batch, $\tau$ is a learnable temperature, and $m_j^{P}$, $m_j^{I}, m_j^{T}$ denote normalized projected features of the mixed features of point clouds, images, and text respectfully. Because the image encoder and text encoder are frozen, we extract and cache the features before training for acceleration. 

In the second stage, We initialize a new trainable 3D encoder. All Feature Mixing Encoders (FM-Encoders) remain frozen in this stage. Then we introduce a mixed point cloud input to further refine the 3D feature representations. Two input point clouds are selected and processed using farthest point sampling (FPS) and point-level mixing to create a novel mixed point cloud. The mixed point cloud is input to the new trainable 3D encoder to obtain mixed 3D feature embeddings. Simultaneously, the frozen Feature Mixing Encoders (FM-Encoders), are used to extract mixed features from their respective inputs. Using a contrastive learning objective, the 3D features of the mixed point cloud are aligned with the mixed features from the frozen encoders, ensuring that the new 3D encoder learns robust and discriminative mixed feature representations from different modalities. The Stage 2 contrastive loss $L^{S2}$ is calculated as:

\begin{small}
\begin{equation}
\begin{split}
L^{S2} = -\frac{1}{6n} \sum_{i} (
& F(m_i^{M}, m_i^{I})
+ F(m_i^{I}, m_i^{M}) + \\
& F(m_i^{M}, m_i^{T}) + F(m_i^{T}, m_i^{M})+ \\[2mm]
& F(m_i^{M}, m_i^{P}) + F(m_i^{P}, m_i^{M})
),
\end{split}
\end{equation}
\end{small}where $m_j^{M}$ denotes normalized projected features of the mixed point clouds $M$. 

By leveraging these two stages, the MM-Mixing training pipeline fully exploits the complementary advantages of image and text encoders, integrating multi-modal information to develop a 3D encoder capable of producing highly discriminative features. In the first stage, the point cloud-image-text feature-level mixing ensures the consistency of augmented features across different modalities, facilitating the 3D encoder's cross-modal understanding. The second stage introduces input-level mixing, providing a vast array of complex and realistic samples that enhance the 3D encoder's generalization ability. Under the constraints of contrastive learning, MM-Mixing maintains the consistency between the features of the mixed point clouds and the mixed features of the point clouds. 
\section{Experiments}
\begin{table*}[]
\centering
\caption{\textbf{Zero-shot 3D classification on ModelNet40, ScanObjectNN and Objaverse-LVIS.} We report the top-1, top-3 and top-5 classification accuracy (\%) for different 3D backbones pre-trained on ShapeNet and Ensembled.}
\resizebox{\linewidth}{!}{
\begin{tabular}{cc|c|ccc|ccc|ccc}
\toprule
\multirow{2}{*}{\begin{tabular}[c]{@{}c@{}}Pre-training\\ Dataset\end{tabular}} & \multirow{2}{*}{\begin{tabular}[c]{@{}c@{}}3D\\ Backbone\end{tabular}} & \multirow{2}{*}{Pre-training Method} & \multicolumn{3}{c|}{ModelNet40} & \multicolumn{3}{c|}{ScanObjectNN} & \multicolumn{3}{c}{Objaverse} \\
 &  &  & Top1 & Top3 & Top5 & Top1 & Top3 & Top5 & Top1 & Top3 & Top5 \\ 
\cmidrule{1-12}
\multirow{2}{*}{\begin{tabular}[c]{@{}c@{}}Projected \\ images \end{tabular}} & \multirow{2}{*}{-} & PointCLIP~\citep{zhang2022pointclip} & 19.3 & 28.6 & 34.8 & 10.5 & 20.8 & 30.6 & 1.9 & 4.1 & 5.8 \\
 &  & PointCLIP v2~\citep{zhu2023pointclip} & 63.6 & 77.9 & 85.0 & 42.2 & 63.3 & 74.5 & 4.7 & 9.5 & 12.9 \\ 
\cmidrule{1-12}
\multirow{10}{*}{ShapeNet} & \multirow{3}{*}{Transformer} & ReCon~\citep{qi2023contrast} & 61.2 & 73.9 & 78.1 & 42.3 & 62.5 & 75.6 & 1.1 & 2.7 & 3.7 \\
 & & CG3D~\citep{hegde2023clip} & 48.7 & 60.7 & 66.5 & 42.5 & 57.3 & 60.8 & 5.0 & 9.5 & 11.6 \\
 & & CLIP2Point~\citep{huang2023clip2point} & 49.5 & 71.3 & 81.2 & 25.5 & 44.6 & 59.4 & 2.7 & 5.8 & 7.9 \\ 
\cmidrule{2-12}
 & \multirow{3}{*}{SparseConv} & OpenShape~\citep{liu2024openshape} & 72.9 & 87.2 & 89.5 & 52.7 & 72.7 & 83.6 & 11.6 & 21.8 & 27.1 \\
 &  & \textbf{MM-Mixing (Ours)} & 75.2 & 88.9 & 91.9 & 60.7 & 79.0 & 87.3 & 13.0 & 23.4 & 28.6 \\
 &  & \textcolor{blue}{↑ \textit{Improve}} & \textcolor{blue}{+2.3} & \blue{+1.7} & \blue{+2.4} & \blue{+8.0} & \blue{+6.3} & \blue{+3.7} & \blue{+1.4} & \blue{+1.6} & \blue{+1.5} \\ 
\cmidrule{2-12}
 & \multirow{4}{*}{Point-BERT} & ULIP~\citep{xue2023ulip} & 60.4 & 79.0 & 84.4 & 51.5 & 71.1 & 80.2 & 6.2 & 13.6 & 17.9 \\
 &  & OpenShape~\citep{liu2024openshape} & 70.3 & 86.9 & 91.3 & 51.3 & 69.4 & 78.4 & 10.8 & 20.2 & 25.0 \\
 &  & \textbf{MM-Mixing (Ours)} & 74.1 & 88.8 & 91.6 & \textbf{61.9} & \textbf{83.0} & \textbf{91.8} & 13.0 & 22.9 & 27.9 \\
 &  & \textcolor{blue}{↑ \textit{Improve}} & \blue{+3.8} & \blue{+1.9} & \blue{+0.3} & \blue{+10.6} & \blue{+13.6} & \blue{+13.4} & \blue{+2.2} & \blue{+2.7} & \blue{+2.9} \\
\cmidrule{1-12}
\multirow{7}{*}{Ensembled} & \multirow{3}{*}{SparseConv} & OpenShape~\citep{liu2024openshape} & 83.4 & 95.6 & 97.8 & 56.7 & 78.9 & 88.6 & 43.4 & 64.8 & 72.4 \\
 &  & \textbf{MM-Mixing (Ours)} & \textbf{86.7} & \textbf{97.7} & \textbf{98.7} & 58.4 & 79.5 & 89.4 & 46.2 & 68.2 & 75.8 \\
 &  & \textcolor{blue}{↑ \textit{Improve}} & \blue{+3.3} & \blue{+2.1} & \blue{+0.9} & \blue{+1.7} & \blue{+0.6} & \blue{+0.8} & \blue{+2.8} & \blue{+3.4} & \blue{+3.4} \\
\cmidrule{2-12}
 & \multirow{4}{*}{Point-BERT} & ULIP~\citep{xue2023ulip} & 75.1 & 88.1 & 93.2 & 51.6 & 72.5 & 82.3 & 26.8 & 44.8 & 52.6 \\ 
 &  & OpenShape~\citep{liu2024openshape} & 84.4 & 96.5 & 98.0 & 52.2 & 79.7 & 88.7 & 46.8 & 69.1 & 77.0 \\
 &  & \textbf{MM-Mixing (Ours)} & 86.0 & 96.6 & 98.4 & 54.3 & 79.9 & 89.1 & \textbf{51.4} & \textbf{73.1} & \textbf{80.1} \\
 &  & \textcolor{blue}{↑\textit{ Improve}} & \blue{+1.6} & \blue{+0.1} & \blue{+0.4} & \blue{+2.1} & \blue{+0.2} & \blue{+0.4} & \blue{+4.6} & \blue{+4.0} & \blue{+3.1} \\
\bottomrule
\end{tabular}
}
\label{table:main}
\vspace{-6pt}
\end{table*}

\begin{table}[]
\centering
\caption{\textbf{Linear probing 3D classification results.} We report the classification accuracy (\%) of Point-BERT on ModelNet40 and three splits of ScanObjectNN.}
\small
\resizebox{\linewidth}{!}{
\begin{tabular}{c|c|c|ccc}
\toprule
\multirow{2}{*}{\begin{tabular}[c]{@{}c@{}}Pre-training\\ Dataset\end{tabular}} & \multirow{2}{*}{\begin{tabular}[c]{@{}c@{}}Pre-training\\ Method\end{tabular}}{} & \multirow{2}{*}{M-40} & \multicolumn{3}{c}{ScanObjectNN} \\
 &  &  & \scriptsize{OBJ-BG} & \scriptsize{OBJ-ONLY} & \scriptsize{PB-T50-RS} \\ 
\midrule
 \multirow{4}{*}{ShapeNet} & ULIP & \textbf{90.6} & 75.4 & 75.4 & 64.8 \\
 & OpenShape & 88.5 & 77.8 & 78.5 & 64.1 \\
 & \textbf{MM-Mixing} & \textbf{90.6} & \textbf{83.3} & \textbf{85.0} & \textbf{73.2} \\
 & \textcolor{blue}{↑ \textit{Improve}} & \blue{+2.1} & \blue{+5.5} & \blue{+6.5} & \blue{+9.1} \\ 
\midrule
 \multirow{3}{*}{Ensembled} & OpenShape & 91.3 & 85.9 & 85.4 & 78.0 \\
 & \textbf{MM-Mixing} & \textbf{91.7} & \textbf{86.9} & \textbf{86.2} & \textbf{79.3} \\
 & \textcolor{blue}{↑ \textit{Improve}} & \blue{+0.4} & \blue{+1.0} & \blue{+0.8} & \blue{+1.3} \\ 
\bottomrule
\end{tabular}
}
\label{table:comparison}
\vspace{-2pt}
\end{table}

\subsection{Experimental Setup}
\noindent\textbf{Pre-training datasets.} In our experimental setup, we utilize datasets following the approach outlined by the state-of-the-art OpenShape~\citep{liu2024openshape}. Our model is pre-trained using triplets generated from four key datasets: ShapeNetCore~\citep{chang2015shapenet}, 3D-FUTURE~\citep{fu20213d}, ABO~\citep{collins2022abo}, and Objaverse~\citep{deitke2023objaverse}. Specifically, the "ShapeNet" training set is composed entirely of triplets from the ShapeNetCore dataset, which includes 52,470 3D shapes along with their associated images and text descriptions. The comprehensive "Ensembled" dataset includes a total of 875,665 triplets, encompassing data from all four datasets, thereby providing a rich source of varied 3D shapes and their corresponding images and texts.

\noindent\textbf{Evaluation datasets.}
For the evaluation of our model, we use a set of datasets that ensures a thorough assessment across different types of 3D data. The Objaverse-LVIS dataset~\citep{deitke2023objaverse}, which is part of our evaluation, contains an extensive variety of categories with 46,832 high-quality shapes distributed across 1,156 LVIS~\citep{gupta2019lvis} categories, offering a diverse and challenging environment for testing. Additionally, we include ModelNet40~\citep{wu20153d} in our evaluation process, a well-known synthetic indoor 3D dataset consisting of 40 categories with a test split of 2,468 shapes. The ScanObjectNN~\citep{uy2019revisiting} dataset, which includes scanned objects from 15 common categories, provides multiple variants such as OBJ-BG, OBJ-ONLY, and PB-T50-RS, each presenting unique challenges~\citep{qi2023contrast, wu2022point}. 
Our experiments are conducted across several distinct tasks: zero-shot 3D classification, linear probing 3D classification, and cross-modal 3D shape retrieval to highlight the capabilities and versatility of our model. Further details regarding the implementation specifics for pre-training and evaluation are provided in the Appendix.

\subsection{Zero-shot 3D Classification}
Zero-shot classification refers to the process where a pre-trained model is directly employed to classify a target dataset without any supervision or prior knowledge from that specific dataset. This task presents a considerable challenge for the model, requiring it to exhibit robust knowledge generalization, deep understanding of 3D shapes, and efficient cross-modal alignment.
We conduct extensive experiments to validate the effectiveness and robustness of our proposed MM-Mixing on three benchmark datasets: ModelNet40, ScanObjectNN, and Objaverse. 

As shown in Table \ref{table:main}, MM-Mixing consistently outperforms state-of-the-art methods under the same configurations (e.g., pre-trained datasets, training epochs, 3D backbones) and enhances the performance of various 3D models across all datasets. For instance, when pre-trained on ShapeNet, MM-Mixing boosts the accuracy of Point-BERT from 51.3\% to 61.9\% on the real-world dataset ScanObjectNN, even surpassing the 52.2\% achieved by OpenShape pre-training on the Ensembled dataset. It indicates that MM-Mixing makes full use of limited multi-modal data to improve the model's understanding of 3D shapes and shows strong performance in handling complex noise interference.

Moreover, on the challenging long-tail dataset, Objaverse, Point-BERT pre-trained with MM-Mixing achieves the accuracy of 51.4\%, outperforming OpenShape's 46.8\%. Another 3D backbone, SparseConv, also showed a 2.8\% improvement in accuracy with our pre-training method. It indicates that existing 3D encoders can be easily incorporated into MM-Mixing framework, leading to a significant enhancement in 3D shape understanding.

When the pre-training data is expanded from ShapeNet to a larger Ensembled dataset, the performance gains from MM-Mixing are slightly diminished. However, it still provides consistent accuracy gains to the models, underscoring the effectiveness of MM-Mixing on large-scale datasets.

\subsection{Linear Probing 3D Classification}
To better adapt the model to the specific classification of downstream tasks, we train a dataset-dependent learnable linear layer to process the 3D features generated by the pre-trained model. Since only the linear layer is activated in this process, the training is lightweight.

The linear probing results are illustrated in Table \ref{table:comparison}. When pre-trained on ShapeNet, MM-Mixing achieves 90.6\% accuracy on ModelNet40, outperforming OpenShape by 2.1\%. On ScanObjectNN, MM-Mixing shows significant improvements, surpassing OpenShape~\citep{liu2024openshape} by 5.5\%, 6.5\% and 9.1\% on OBJ-BG, OBJ-ONLY, and PB-T50-RS, respectively.
When using the Ensembled dataset for pre-training, MM-Mixing maintains its lead with 91.7\% accuracy on ModelNet40 and consistent superiority on ScanObjectNN three subsets, with accuracies of 86.9\%, 86.2\%, and 79.3\% respectively. These findings emphasize that MM-Mixing has learned robust and discriminative 3D feature representations during pre-training, which can be efficiently applied to downstream specific classification tasks through a simple linear layer.


\begin{table}[]
\centering
\caption{\textbf{Ablation studies on Mixing level in alignment.} “FM” represents feature-level mixing. “IM” represents input-level mixing.}
\small

\resizebox{\linewidth}{0.55in}{
\begin{tabular}{c|cc|cc|cc}
\toprule
\multirow{2}{*}{\begin{tabular}[c]{@{}c@{}}Mixing\\ level\end{tabular}} &  \multicolumn{2}{c|}{ModelNet40} & \multicolumn{2}{c|}{ScanObjectNN} & \multicolumn{2}{c}{Objaverse} \\ 
 & Top1 & Top5 & Top1 & Top5 & Top1 & Top5 \\ 
 \midrule
Baseline & 72.9 & 89.5 & 52.7 & 83.6 & 11.6 & 27.1 \\
FM & 74.1 & 90.1 & 56.4 & 84.7 & 12.2 & 27.3 \\
IM & 73.8 & 90.4 & 58.9 & 85.2 & 12.4 & 27.5 \\
FM+IM & \textbf{75.2} & \textbf{91.9} & \textbf{60.7} & \textbf{87.3} & \textbf{13.0} & \textbf{28.6} \\
\bottomrule
\end{tabular}

}
\label{table: level}
\vspace{-0pt}
\end{table}

\begin{table}[]
\centering
\caption{\textbf{Ablation studies on Alignment stage.} “One stage” represents all learnable networks are trained simultaneously.}
\small
\resizebox{\linewidth}{!}{
\begin{tabular}{c|cc|cc|cc}
\toprule
\multirow{2}{*}{Stage} &  \multicolumn{2}{c|}{ModelNet40} & \multicolumn{2}{c|}{ScanObjectNN} & \multicolumn{2}{c}{Objaverse} \\ 
 & Top1 & Top5 & Top1 & Top5 & Top1 & Top5 \\ 
 \midrule
One stage & 73.6 & 90.2 & 59.5 & 85.8 & 12.3 & 27.7 \\
Two stages & \textbf{75.2} & \textbf{91.9} & \textbf{60.7} & \textbf{87.3} & \textbf{13.0} & \textbf{28.6} \\
\bottomrule
\end{tabular}
}
\label{table: stage}
\vspace{-2pt}
\end{table}

\subsection{Ablation Study}
We systematically study the impact of different components in MM-Mixing on the model's performance, including the mixing level, alignment stage, modality loss function, and training costs analysis. All results are the classification accuracy (\%) of SparseConv pre-trained on ShapeNet.

\noindent\textbf{Mixing levels in alignment.} 
 We investigate the impact of different mixing levels, including Feature-level Mixing (FM), Input-level Mixing (IM), and their combination (FM+IM). Compared to the baseline without mixing, all three strategies consistently improve the performance across all datasets. In Table \ref{table: level}, Feature-level Mixing (FM) and Input-level Mixing (IM) individually contribute to the performance gains, and their combination (FM+IM) further improves the results. It confirms that the two mixing levels complement each other: Feature-level Mixing (FM) ensures cross-modal consistency in the feature latent space, while Input-level Mixing (IM) refines the realistic point cloud representation with challenging samples. Together, they enhance the model's ability of 3D understanding.
 
\noindent\textbf{Alignment stages.} As shown in Table \ref{table: stage}, we evaluate the effectiveness of our two-stage alignment design. Feature-level Mixing(FM) is first employed to align 3D features with their corresponding modalities. In the second stage, mixed point cloud inputs are introduced to further align the four kinds of mixed features across the three modalities. The other approach is to align the mixed features of two levels simultaneously in one stage. Compared to one-stage alignment, the two-stage alignment method can better utilize diverse mixed samples to enhance cross-modal consistency.

\noindent\textbf{Modality loss functions.} Our ablation studies on different modality loss functions are shown in Table \ref{table: loss}. The text loss $\mathcal{L_T}$ provides a strong foundation for learning 3D representations with semantic information, while the image loss $\mathcal{L_I}$ and point cloud loss $\mathcal{L_P}$ offer complementary visual and shape information, enhancing the model's performance. The combination of all three modality loss functions consistently achieves the best results across all datasets, demonstrating the effectiveness of our framework.

\begin{table}[]
\centering
\caption{\textbf{Ablation studies on Modality loss function.} $\mathcal{L_T}$ represents the text loss. $\mathcal{L_I}$ represents the image loss. $\mathcal{L_P}$ represents the point cloud loss.}
\small
\resizebox{\linewidth}{0.55in}{
\begin{tabular}{ccc|cc|cc|cc}
\toprule
\multirow{2}{*}{$\mathcal{L_T}$} & 
\multirow{2}{*}{$\mathcal{L_I}$} & 
\multirow{2}{*}{$\mathcal{L_P}$} & 
\multicolumn{2}{c|}{ModelNet40} & \multicolumn{2}{c|}{ScanObjectNN} & \multicolumn{2}{c}{Objaverse} \\ 
 & & & Top1 & Top5 & Top1 & Top5 & Top1 & Top5 \\ 
\midrule
\checkmark &  &  & 72.6 & 89.2 & 58.9& 85.4 & 11.4 & 24.7 \\
\checkmark & \checkmark &  & 73.8 & 90.8 & \textbf{60.7} & 85.4 & 12.5 & 27.9 \\
\checkmark &  & \checkmark & 73.9 & 89.7 & 60.4 & 85.6 & 11.7& 25.7 \\
\checkmark & \checkmark & \checkmark & \textbf{75.2} & \textbf{91.9} & \textbf{60.7} & \textbf{87.3} & \textbf{13.0} & \textbf{28.6} \\ \bottomrule
\end{tabular}
}
\label{table: loss}
\vspace{-2pt}
\end{table}

\noindent\textbf{Training costs analysis}. Notably, the epochs of one-stage methods are the same as the two-stage training epochs of MM-Mixing for a fair comparison. Both 3D encoders are trained independently for the duration of one stage without shared weights. The experimental results demonstrate that the performance gains of MM-Mixing primarily stem from our mixing-based alignment framework, and the two-stage training framework further enhances the effectiveness of dual-level mixing. Moreover, for previous methods like OpenShape, adding additional training costs (e.g. training time and training parameters) does not significantly improve the performance of the 3D backbone (See Appendix for more details).

\subsection{Qualitative Analysis}

\begin{figure*}[t]
    \centering
        \includegraphics[width=\linewidth]{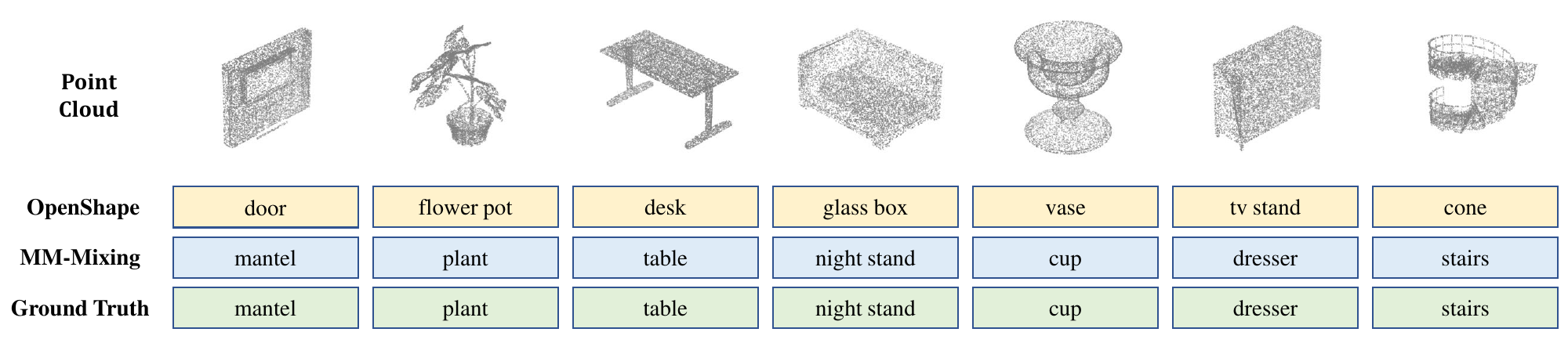}
    \caption{\textbf{Hard sample recognition on ModelNet40.} Compared to OpenShape, MM-Mixing enables the model to better capture typical features across different categories and the ability to distinguish hard samples.}
    \label{fig: classification}
    \vspace{-2pt}
\end{figure*}

\begin{figure*}[t]
    \centering
        \includegraphics[width=\linewidth]{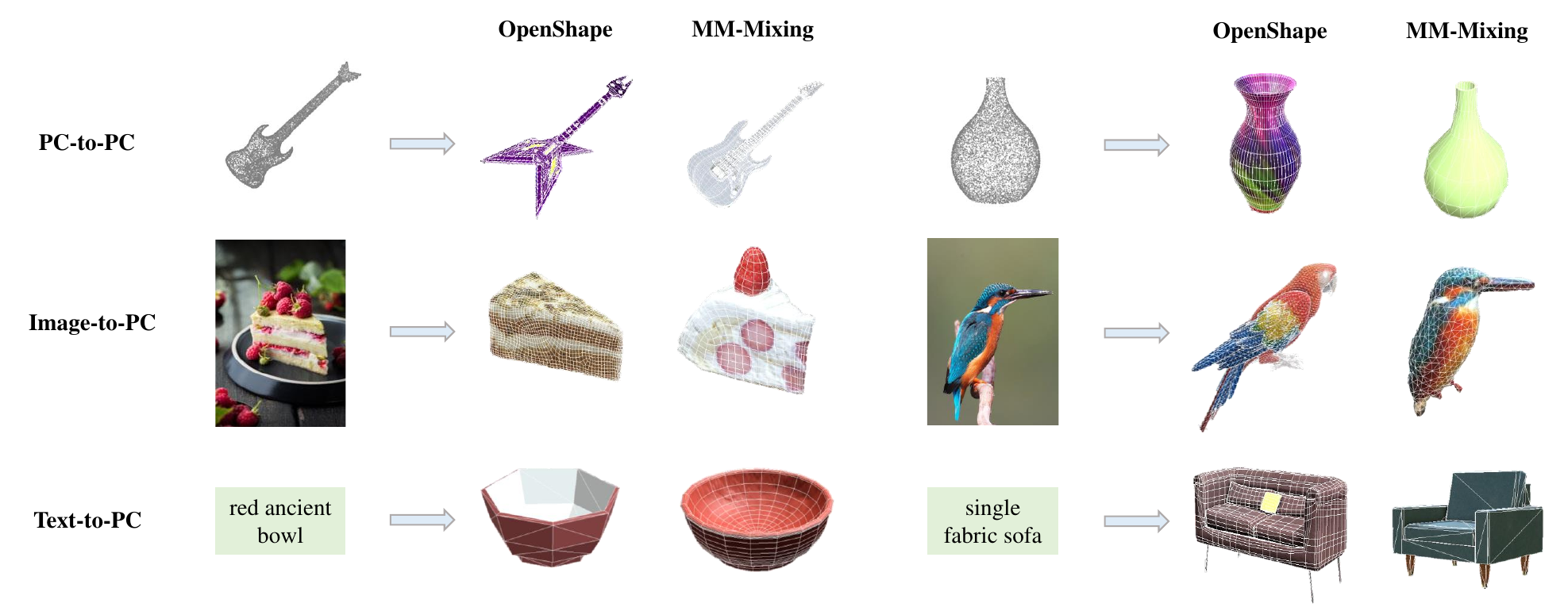}
    \caption{\textbf{Cross-modal 3D shape retrieval on Objaverse.} Compared to OpenShape, MM-Mixing enhances the model's understanding of point cloud shapes, image colors, and textual descriptions, effectively improving cross-modal 3D shape retrieval capabilities. PC represents Point Cloud.}
    \label{fig: retrieval}
    \vspace{-6pt}
\end{figure*}

\noindent\textbf{Hard sample recognition.} 
In real-world scenarios, numerous objects exhibit similar morphological or visual characteristics despite belonging to distinct categories. We designate these challenging instances as "hard samples." There are some such category pairs in ModelNet40, such as: "vase \& cup", "table \& desk", "TV stand \& dresser", and "plant \& flower pot". As illustrated in Figure \ref{fig: classification}, MM-Mixing demonstrates the capability to capture subtle differences between objects that may appear similar but have different categories. For instance, MM-Mixing can distinguish between cups and vases by accurately understanding the correspondence between the appearance and function of the objects. Additionally, it can leverage detailed features (e.g. the presence of a drawer) to prevent misidentifying a table as a desk. It can be confirmed that MM-Mixing enhances model performance in 3D object recognition, particularly in scenarios with confusing samples and noise interference.

\noindent\textbf{Cross-modal 3D shape retrieval.}
The visualization in Figure \ref{fig: retrieval} illustrates the superior performance of our method, MM-Mixing, compared to OpenShape in various cross-modal retrieval tasks. For PC-to-PC retrieval, MM-Mixing demonstrates a finer capture of shape details, as seen with the more accurate symmetrical guitar shape. In Image-to-PC retrieval, our method excels in preserving color details, which can retrieve more rational and approximate point clouds, such as the cake example. Additionally, in text-to-PC retrieval, MM-Mixing shows enhanced compatibility with complex textual descriptions, accurately reflecting shape, color, and material details, as evidenced by the "single fabric sofa" example. These results highlight MM-Mixing's effectiveness in improving shape fidelity, color accuracy, and textual comprehension in cross-modal retrieval.

\section{Conclusion}
In this paper, we propose \textbf{MM-Mixing}, a multimodal mixing alignment approach that addresses the challenges of multi-modal alignment and enhances model generation for 3D understanding. By integrating the mixing-based method with multimodal data through a two-stage training pipeline, MM-Mixing enhances the performance and generalization capabilities of the models, which ensures a cohesive enhancement of features from different modalities. Extensive experiments demonstrate the effectiveness of MM-Mixing, significantly boosting baseline performance across various settings, including zero-shot 3D classification, linear probing 3D classification, and cross-modal 3D shape retrieval.
Moreover, MM-Mixing addresses the previously unexplored issue of multimodal mixing alignment, offering a simple yet effective solution that can be easily integrated into existing frameworks. As 3D vision continues to evolve and find applications in various domains, MM-Mixing represents a significant step forward in meeting the challenges of robust and generalizable models.
Our methodology will contribute to further advancements in the field, supporting the ongoing evolution of 3D understanding within multimodal learning.

\bibliography{aaai25}
\section{Appendix}
\subsection{Training Details}
Our training setup utilizes four A100 GPUs, with each training batch consisting of 200 examples. In alignment with the methodologies employed by OpenShape~\cite{liu2024openshape}, we enhance the efficiency of our training process by precaching the CLIP~\cite{radford2021learning} embeddings for both text and images corresponding to all the shapes. This optimization significantly speeds up the training, enabling the model to converge in approximately 400 A100 GPU hours. We use the AdamW optimizer~\cite{loshchilov2017decoupled} and adopt an exponential learning rate schedule and conduct a range test to determine the optimal initial learning rate. Specifically, we set the learning rate at 5e-4 for Point-BERT~\cite{yu2022point} and at 1e-3 for all other models.

\subsection{Experiment Results}
\textbf{Reviewing pre-training cost.} It is generally assumed that a simple two-stage pre-training framework will inherently bring performance gains. In this context, the contribution of the dual-level mixed alignment method might be questioned. 
To address this, we conducted a comprehensive analysis of the relationship between training costs and model performance, focusing on pre-trained model parameters and pre-training epoch, as shown in Table \ref{table: cost}.
Our findings demonstrate that MM-Mixing consistently outperforms OpenShape across all datasets, irrespective of the pre-training configuration. Notably, when pre-training epochs are set to 1000, MM-Mixing achieves superior performance with only 41.3M parameters compared to OpenShape, which requires twice the number of parameters. This suggests that the core of MM-Mixing's enhanced 3D understanding capability lies in its mixed-based alignment method, a feature absent in OpenShape.
Furthermore, our results indicate that the two-stage pre-training approach yields more substantial performance gains for MM-Mixing compared to the single-stage framework. This improvement can be attributed to the enhanced consistency of the dual-level mixing process. These findings underscore the significance of our proposed method in advancing 3D understanding capabilities.


\begin{table}[]
\centering
\caption{\textbf{The impact of pre-training cost.} We report the classification accuracy (\%) of Point-BERT pre-trained on ShapeNet.}
\small
\scalebox{0.65}{
\begin{tabular}{ccc|cc|cc|cc}
\toprule
\multirow{2}{*}{\begin{tabular}[c]{@{}c@{}}Pre-Training\\ Param. (M)\end{tabular}} & \multirow{2}{*}{\begin{tabular}[c]{@{}c@{}}Pre-Training\\ Epoch\end{tabular}} & \multirow{2}{*}{\begin{tabular}[c]{@{}c@{}}Pre-Training \\ Method\end{tabular}} & \multicolumn{2}{c|}{ModelNet40} & \multicolumn{2}{c|}{ScanObjectNN} & \multicolumn{2}{c}{Objaverse} \\ 
 & & & Top1 & Top5 & Top1 & Top5 & Top1 & Top5 \\ 
\midrule
\multirow{6}{*}{41.3} & \multirow{3}{*}{500} & OpenShape & 72.2 & 89.1 & 52.4 & 82.3 & 10.8 & 26.0 \\ 
 & & MM-Mixing & 72.5 & 91.2 & 58.3 & 83.4 & 11.7 & 26.7 \\
 & & \textcolor{blue}{↑ \textit{Improve}} & \blue{+0.3} & \blue{+2.1} & \blue{+5.9} & \blue{+1.1} & \blue{+0.9} & \blue{+0.7} \\
\cmidrule{2-9}
 & \multirow{3}{*}{1000} & OpenShape & 72.9 & 89.5 & 52.7 & 83.6 & 11.6 & 27.1 \\
 & & MM-Mixing & 73.6 & 90.2 & 59.5 & 85.8 & 12.3 & 27.7 \\
 & & \textcolor{blue}{↑ \textit{Improve}} & \blue{+0.7} & \blue{+0.7} & \blue{+6.8} & \blue{+2.2} & \blue{+0.7} & \blue{+0.6} \\
\midrule
\multirow{6}{*}{82.6} & \multirow{3}{*}{500} & OpenShape & 73.0 & 89.1 & 52.6 & 84.5 & 11.4 & 27.2 \\ 
 & & MM-Mixing & 74.4 & 91.6 & 60.6 & 87.0 & 12.7 & 28.4 \\
 & & \textcolor{blue}{↑ \textit{Improve}} & \blue{+1.4} & \blue{+2.5} & \blue{+8.0} & \blue{+2.5} & \blue{+1.3} & \blue{+1.2} \\
\cmidrule{2-9}
 & \multirow{3}{*}{1000} & OpenShape & 73.2 & 89.4 & 53.1 & 83.9 & 11.8 & 27.4 \\
 & & MM-Mixing & 75.2 & 91.9 & 60.7 & 87.3 & 13.0 & 28.6 \\
 & & \textcolor{blue}{↑ \textit{Improve}} & \blue{+2.0} & \blue{+2.5} & \blue{+7.6} & \blue{+3.4} & \blue{+1.2} & \blue{+1.2} \\
\bottomrule
\end{tabular}
}
\label{table: cost}
\vspace{-12pt}
\end{table}

\begin{table*}[]
\centering
\caption{\textbf{Zero-shot 3D classification results by category on real-world ScanObjectNN dataset.} We report the classification accuracy(\%) of each category and the mean accuracy of all categories.}
\resizebox{\linewidth}{!}{
\begin{tabular}{c|c|c|cccccccccccccccc}
\toprule
 & Model & Aug & Avg & bag & bin & box & cabinet & chair & desk & display & door & shelf & table & bed & pillow & sink & sofa & toilet \\ \cmidrule{1-19}
\multirow{6}{*}{Top1} & \multirow{3}{*}{SparseConv} & OpenShape & 51.4 & 58.4 & 20.9 & 11.9 & 0.0 & 90.9 & 61.1 & 51.9 & 94.1 & 63.7 & 51.4 & 57.0 & 41.0 & 51.7 & 46.0 & 72.0 \\
 &  & MM-Mixing & 62.0 & 68.8 & 39.8 & 30.8 & 0.6 & 87.1 & 76.5 & 77.4 & 99.1 & 64.0 & 66.8 & 70.4 & 70.5 & 60.2 & 47.6 & 70.7 \\
 &  & \textcolor{blue}{↑\textit{improve}} & \blue{10.6} & \blue{10.4} & \blue{18.9} & \blue{18.9} & \blue{0.6} & \blue{-3.8} & \blue{15.4} & \blue{25.5} & \blue{5.0} & \blue{0.3} & \blue{15.4} & \blue{13.4} & \blue{29.5} & \blue{8.5} & \blue{1.6} & \blue{-1.3} \\ \cmidrule{2-19} 
 & \multirow{3}{*}{Point-BERT} & Openshape & 48.5 & 53.2 & 11.4 & 18.8 & 1.4 & 83.0 & 74.5 & 64.6 & 93.7 & 58.4 & 49.8 & 58.5 & 12.5 & 44.9 & 43.5 & 58.5 \\
 &  & MM-Mixing & 61.6 & 53.3 & 32.2 & 43.6 & 4.6 & 92.2 & 81.9 & 76.8 & 97.3 & 76.8 & 51.5 & 72.6 & 65.7 & 48.3 & 59.5 & 67.8 \\
 &  & \textcolor{blue}{↑\textit{improve}} & \blue{13.1} & \blue{0.1} & \blue{20.8} & \blue{24.8} & \blue{3.2} & \blue{9.2} & \blue{7.4} & \blue{12.2} & \blue{3.6} & \blue{18.4} & \blue{1.7} & \blue{14.1} & \blue{53.2} & \blue{3.4} & \blue{16.0} & \blue{9.3} \\ \cmidrule{1-19}
\multirow{6}{*}{Top3} & \multirow{3}{*}{SparseConv} & Openshape & 73.9 & 84.4 & 54.7 & 47.8 & 9.8 & 95.2 & 85.2 & 88.4 & 98.2 & 85.0 & 87.5 & 74.0 & 60.0 & 74.6 & 73.6 & 91.5 \\
 &  & MM-Mixing & 82.7 & 94.8 & 78.6 & 65.8 & 8.9 & 94.4 & 89.3 & 99.5 & 100 & 86.1 & 87.1 & 89.6 & 90.5 & 80.5 & 83.1 & 92.7 \\
 &  & \textcolor{blue}{↑\textit{improve}} & \blue{8.8} & \blue{10.4} & \blue{23.9} & \blue{18.0} & \blue{-0.9} & \blue{-0.8} & \blue{4.1} & \blue{11.1} & \blue{1.8} & \blue{1.1} & \blue{-0.4} & \blue{15.6} & \blue{30.5} & \blue{5.9} & \blue{9.5} & \blue{1.2} \\ \cmidrule{2-19} 
 & \multirow{3}{*}{Point-BERT} & Openshape & 70.1 & 90.9 & 38.3 & 54.7 & 4.0 & 89.3 & 87.9 & 98.9 & 96.8 & 79.4 & 83.8 & 73.3 & 36.1 & 70.3 & 67.7 & 80.4 \\
 &  & MM-Mixing & 82.8 & 92.0 & 57.5 & 84.6 & 28.5 & 96.9 & 87.9 & 97.2 & 99.6 & 95.9 & 84.4 & 89.6 & 81.9 & 71.6 & 90.5 & 83.5 \\
 &  & \textcolor{blue}{↑\textit{improve}} & \blue{12.7} & \blue{1.1} & \blue{19.2} & \blue{29.9} & \blue{24.5} & \blue{7.6} & \blue{0.0} & \blue{-1.7} & \blue{2.8} & \blue{16.5} & \blue{0.6} & \blue{16.3} & \blue{45.8} & \blue{1.3} & \blue{22.8} & \blue{3.1} \\ \cmidrule{1-19}
\multirow{6}{*}{Top5} & \multirow{3}{*}{SparseConv} & Openshape & 84.9 & 93.5 & 77.6 & 70.9 & 32.5 & 97.2 & 91.9 & 99.3 & 99.5 & 92.1 & 94.1 & 83.7 & 76.1 & 83.9 & 86.6 & 93.9 \\
 &  & MM-Mixing & 90.7 & 100.0 & 90.1 & 89.7 & 38.6 & 96.0 & 91.3 & 100.0 & 100.0 & 95.1 & 88 & 97.8 & 98.1 & 86.4 & 90.9 & 98.7 \\
 &  & \textcolor{blue}{↑\textit{improve}} & \blue{5.8} & \blue{6.5} & \blue{12.5} & \blue{18.8} & \blue{6.1} & \blue{-1.2} & \blue{-0.6} & \blue{0.7} & \blue{0.5} & \blue{3.0} & \blue{-6.1} & \blue{14.1} & \blue{22.0} & \blue{2.5} & \blue{4.3} & \blue{4.8} \\ \cmidrule{2-19} 
 & \multirow{3}{*}{Point-BERT} & Openshape & 81.2 & 98.7 & 59.2 & 83.7 & 22.7 & 91.9 & 92.6 & 99.8 & 98.6 & 91.0 & 90 & 80.7 & 55.2 & 83.9 & 83.8 & 86.5 \\
 &  & MM-Mixing & 91.7 & 98.9 & 77.5 & 97.4 & 60.7 & 98.0 & 92.8 & 100.0 & 100.0 & 99.3 & 90.8 & 99.3 & 90.5 & 85.1 & 94.5 & 90.5 \\
 &  & \textcolor{blue}{↑\textit{improve}} & \blue{10.5} & \blue{0.2} & \blue{18.3} & \blue{13.7} & \blue{38.0} & \blue{6.1} & \blue{0.2} & \blue{0.2} & \blue{1.4} & \blue{8.3} & \blue{0.8} & \blue{18.6} & \blue{35.3} & \blue{1.2} & \blue{10.7} & \blue{4.0} \\ \bottomrule
\end{tabular}
}
\label{table:class}
\vspace{-12pt}
\end{table*}

\noindent\textbf{Zero-shot 3D classification on ScanObjectNN.} To further validate the effectiveness of MM-Mixing in enhancing the generalization ability of 3D representation learning models, we conduct zero-shot classification experiments on the real-world ScanObjectNN~\cite{uy2019revisiting} dataset. As shown in Table \ref{table:class}, MM-Mixing significantly improves the performance of both SparseConv\cite{xue2023ulip} and Point-BERT. For SparseConv, MM-Mixing boosts the average accuracy from 51.4\% to 62.0\%, achieving an improvement of 10.6 percentage points. Similarly, for Point-BERT, MM-Mixing enhances the average accuracy from 48.5\% to 61.6\%, resulting in an improvement of 13.1 percentage points.

Notably, MM-Mixing brings improvements in most object categories. For SparseConv, all categories except chair and toilet witness accuracy gains, with the most significant improvements in the display and pillow categories, reaching 25.5 and 29.5 percentage points, respectively. For Point-BERT, all categories except bag experience performance enhancements, with the pillow category showcasing the most remarkable improvement of 53.2 percentage points.
However, some categories remain challenging. For instance, the cabinet category exhibits extremely low accuracy (below 5\%) in all cases, indicating that this category may be particularly difficult to recognize and require further exploration of alternative strategies to boost its performance.
Comparing the two 3D backbones, although Point-BERT initially underperforms SparseConv, MM-Mixing elevates Point-BERT's performance to a level comparable to SparseConv (61.6\% vs. 62.0\%). This observation reinforces the notion that it may be particularly well-suited for Transformer-based models like Point-BERT.
It is worth noting that MM-Mixing leads to performance degradation in a few categories. For example, in SparseConv, the chair and toilet categories experience a drop of 3.8 and 1.3 percentage points, respectively. This suggests that MM-Mixing may have negative impacts on certain categories, warranting further investigation into the underlying reasons and the development of targeted improvement strategies.

\begin{table}[]
\centering
\caption{\textbf{The impact of the number of FC layers.} We report the classification accuracy (\%) of SparseConv and PointBERT on ModelNet40 and three splits of ScanObjectNN.}
\resizebox{\linewidth}{!}{
\begin{tabular}{ccccccc}
\toprule
\multirow{2}{*}{Pre-training Dataset} & \multirow{2}{*}{method} & \multirow{2}{*}{layers} & \multirow{2}{*}{ModelNet40} & \multicolumn{3}{c}{ScanObjectNN} \\
 &  &  &  & OBJ-BG & OBJ-ONLY & PB-T50\_RS \\ \cmidrule{1-7}
\multirow{6}{*}{ShapeNet} & \multirow{3}{*}{SparseConv} & 1 & 90.0 & 83.6 & 85.9 & 74.4 \\
 &  & 2 & 90.3 & \textbf{86.6} & \textbf{87.1} & \textbf{75.5} \\
 &  & 3 & \textbf{90.6} & 85.9 & 86.7 & 75.1 \\ \cmidrule{2-7} 
 & \multirow{3}{*}{Point-BERT} & 1 & 90.6 & 83.3 & 85.0 & 73.2 \\
 &  & 2 & 91.1 & 88.7 & 88.6 & 78.6 \\
 &  & 3 & \textbf{92.0} & \textbf{89.3} & \textbf{89.0} & \textbf{78.4} \\ \cmidrule{1-7}
\multirow{6}{*}{Ensembled} & \multirow{3}{*}{SparseConv} & 1 & 91.5 & 86.6 & 85.6 & 78.7 \\
 &  & 2 & 91.7 & 87.3 & 86.7 & 78.9 \\
 &  & 3 & \textbf{91.8} & \textbf{88.0} & \textbf{87.3} & \textbf{79.0} \\ \cmidrule{2-7} 
 & \multirow{3}{*}{Point-BERT} & 1 & 91.7 & 86.9 & 86.2 & 79.3 \\
 &  & 2 & 92.6 & 88.2 & 88.0 & 81.9 \\
 &  & 3 & \textbf{93.4} & \textbf{90.4} & \textbf{89.3} & \textbf{83.2} \\ \bottomrule
\end{tabular}
}
\label{table:fc}
\vspace{-12pt}
\end{table}

\begin{figure}[t]
    \centering
        \includegraphics[width=\linewidth]{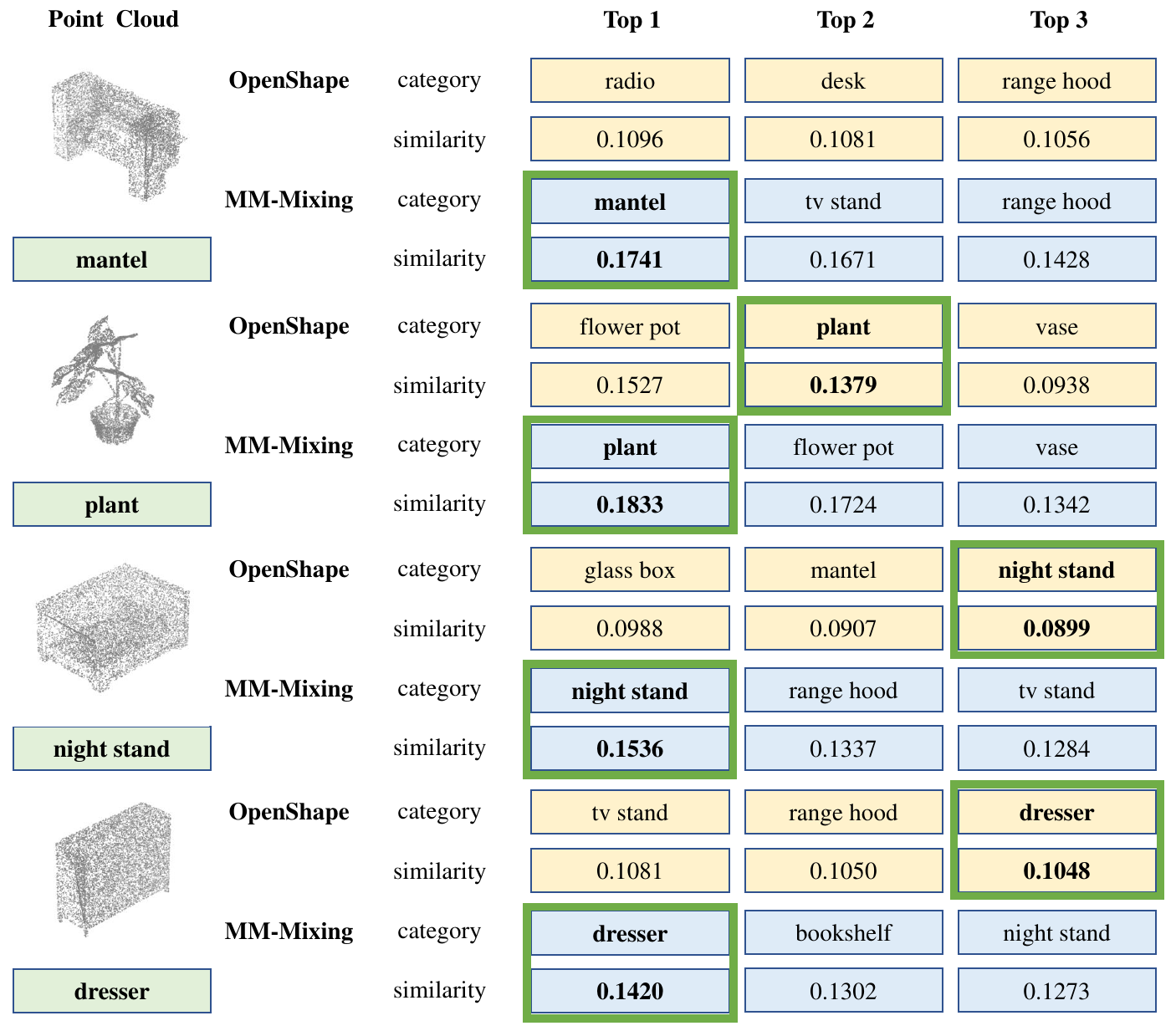}
    \caption{\textbf{Hard sample recognition similarity scores on ModelNet40.} Compared to OpenShape, MM-Mixing not only provides the correct top categoriy, but also obtains higher similarity scores.}
    \label{fig: app-classification}
    \vspace{-12pt}
\end{figure}

\noindent\textbf{The impact of the number of FC layers.}
Table \ref{table:fc} provides a comprehensive analysis of the impact of varying the number of fully connected (FC) layers on the performance of linear probing in different pre-training and evaluation scenarios.

When pre-trained on ShapeNet~\cite{chang2015shapenet}, the SparseConv model shows a progressive improvement in performance on ModelNet40~\cite{wu20153d} and ScanObjectNN datasets as the number of FC layers increases from 1 to 3. Specifically, the optimal performance on ModelNet40 (90.6\%) and ScanObjectNN subsets (OBJ-BG: 86.6\%, OBJ-ONLY: 87.1\%, PB-T50\_RS: 75.5\%) is achieved with two FC layers, indicating that a moderate complexity in the FC layer structure can yield significant gains.
For the Point-BERT model pre-trained on ShapeNet, an increase in the number of FC layers consistently enhances performance across all datasets, with the highest accuracy observed at three layers (ModelNet40: 92.0\%, OBJ-BG: 89.3\%, OBJ-ONLY: 89.0\%, PB-T50\_RS: 78.4\%). This suggests that Point-BERT benefits more substantially from deeper FC layers compared to SparseConv.
In the case of the ensembled pre-training dataset, similar trends are observed. The SparseConv model achieves its best performance with three FC layers (ModelNet40: 91.8\%, OBJ-BG: 88.0\%, OBJ-ONLY: 87.3\%, PB-T50\_RS: 79.0\%), while the Point-BERT model significantly outperforms with three FC layers as well (ModelNet40: 93.4\%, OBJ-BG: 90.4\%, OBJ-ONLY: 89.3\%, PB-T50\_RS: 83.2\%). The results indicate that ensembling pre-training data and increasing the FC layer depth synergistically enhance the model's ability to generalize and accurately classify 3D objects.

Overall, our findings underscore the importance of optimizing the FC layer depth in linear probing to achieve superior model performance, with Point-BERT demonstrating a greater propensity for performance improvement with increased layer depth compared to SparseConv.

\begin{figure}[t]
    \centering
        \includegraphics[width=\linewidth]{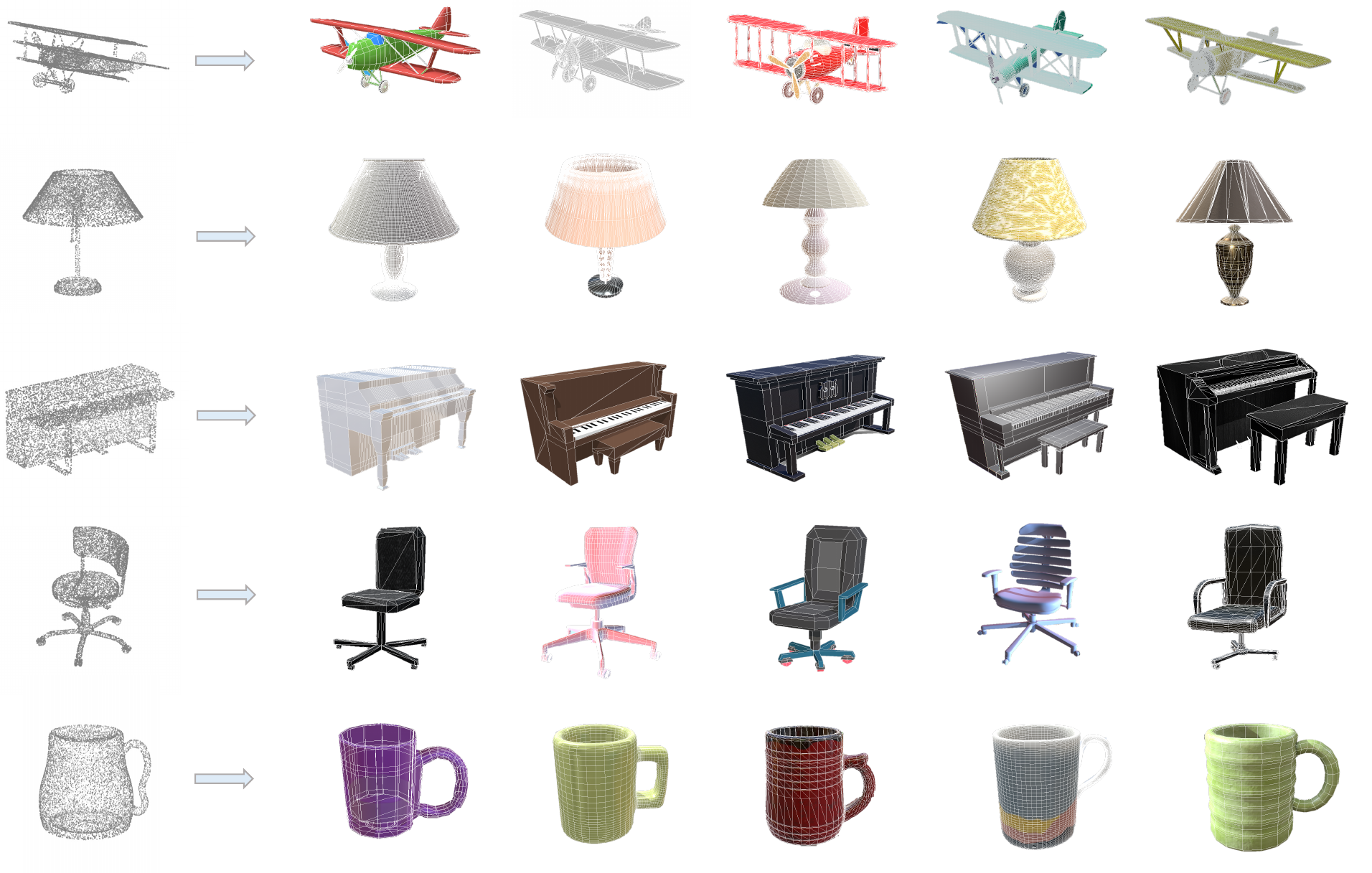}
    \caption{Point Cloud to 3D Shape Retrieval on Objaverse. }
    \label{fig: app-PtoP}
    \vspace{-12pt}
\end{figure}
\begin{figure}[t]
    \centering
        \includegraphics[width=\linewidth]{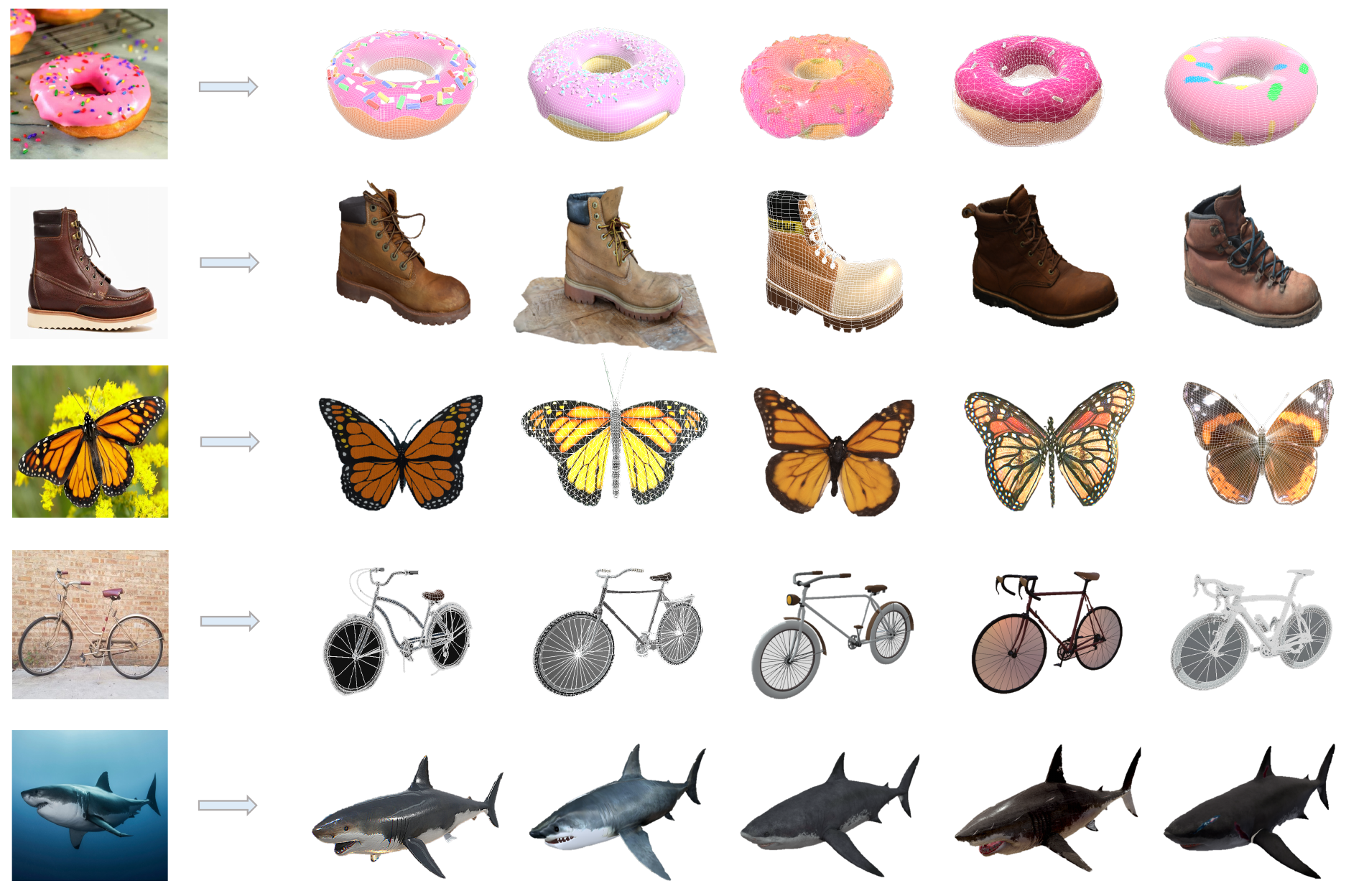}
    \caption{Image to 3D Shape Retrieval on Objaverse.}
    \label{fig: app-ItoP}
    \vspace{-12pt}
\end{figure}
\begin{figure}[t]
    \centering
        \includegraphics[width=\linewidth]{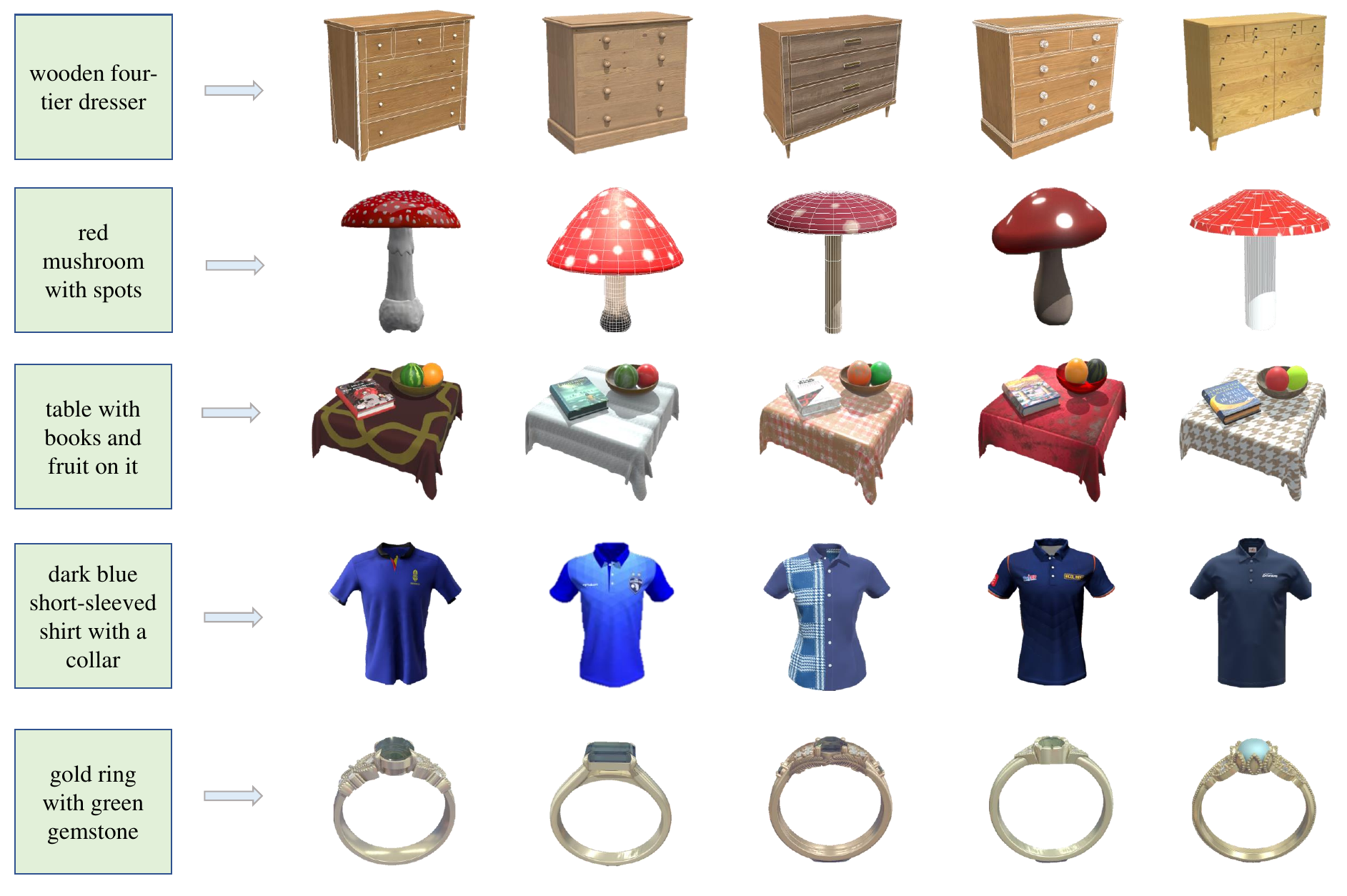}
    \caption{Text to 3D Shape Retrieval on Objaverse.}
    \label{fig: app-TtoP}
    \vspace{-12pt}
\end{figure}

\subsection{Hard Sample Recognition}
Hard sample recognition qualitative results on the ModelNet40 dataset in Figure \ref{fig: app-classification} clearly demonstrate the superior performance of MM-Mixing compared to the previous method, OpenShape. MM-Mixing consistently achieves higher similarity scores and more accurate top predictions across various categories. For instance, in the case of a "mantel," MM-Mixing correctly identifies it as the top category with a similarity score of 0.1741, while OpenShape incorrectly labels it as a "radio." Similar trends are observed for other categories such as "plant", "night stand", and "dresser", where MM-Mixing not only provides the correct top category but also achieves higher similarity scores, indicating a stronger alignment with the true categories. 

These results highlight the robustness and effectiveness of MM-Mixing in accurately classifying point cloud data. Its strong ability to distinguish challenging samples positions it as a more reliable framework for zero-shot 3D classification tasks, unlocking greater potential in practical applications that demand precise 3D shape recognition.

\subsection{Cross-modal Retrieval}
\textbf{Point cloud to 3D shape retrieval.}
Figure \ref{fig: app-PtoP} shows the experimental results on the Objaverse dataset for point cloud to 3D shape retrieval.
As we can see, MM-Mixing successfully matches the input point clouds to their corresponding 3D shapes with high accuracy in most cases, highlighting its advantage in 3D shape understanding. However, in some complex shapes, such as pianos, there is a slight discrepancy in detail accuracy, indicating that while MM-Mixing excels in overall shape matching, there is room for improvement in handling intricate and detailed structures. Overall, MM-Mixing significantly enhances retrieval accuracy, showcasing its potential in accurate 3D shape recognition.

\noindent\textbf{Image to 3D shape retrieval.}
Figure \ref{fig: app-ItoP} shows the experimental results on the Objaverse dataset for image to 3D shape retrieval. The input images, ranging from everyday objects like a donut to more complex items like bicycles and sharks, are effectively represented in the retrieved 3D shapes, which demonstrate the exceptional capability of MM-Mixing in accurately matching 2D images to their 3D counterparts. For instance, the retrieval of the "pink frosted donut with sprinkles" shows meticulous attention to texture and color, which are critical for recognizing food items. Similarly, the retrieval of the "brown boot" captures the detailed design and structure, showcasing our proposed method's proficiency in handling objects with intricate patterns. Therefore, our MM-Mixing effectively bridges the gap between 2D representations and 3D shapes.
\begin{figure*}[t]
    \centering
        \includegraphics[width=\linewidth]{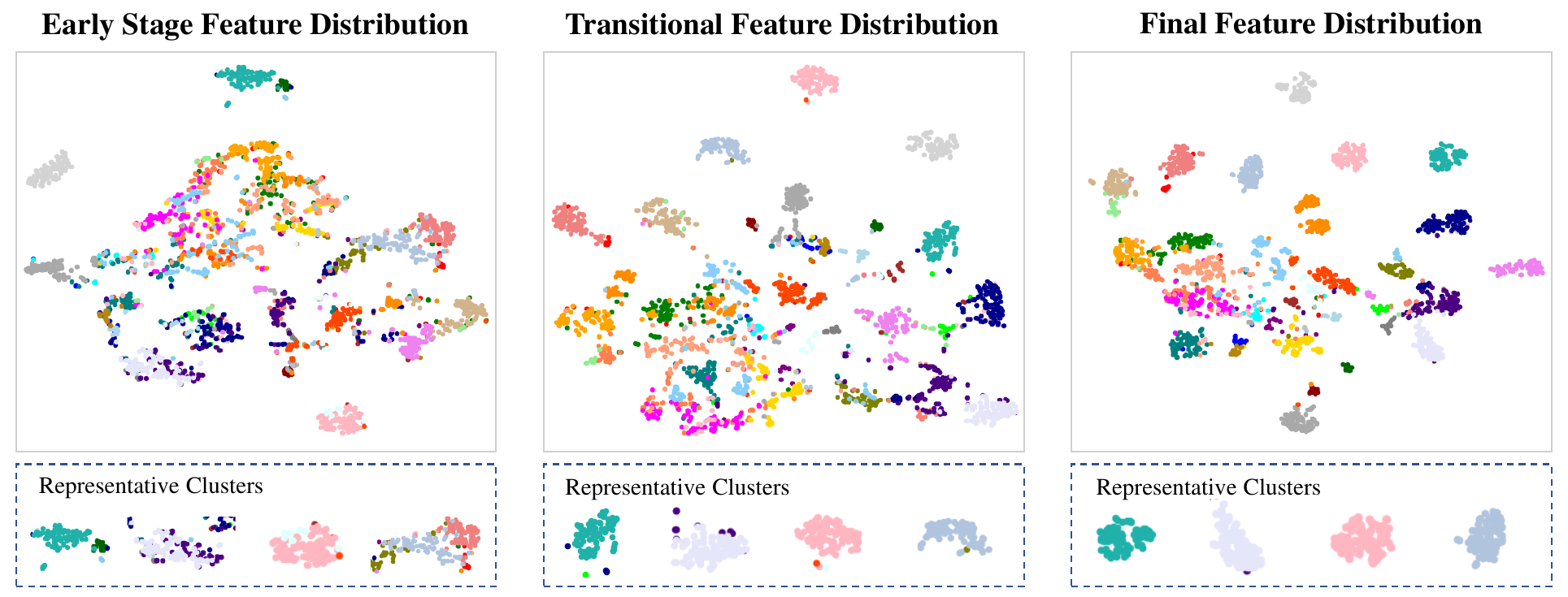}
    \caption{\textbf{Feature distribution visualization on ModelNet40.} Top: An overview of the evolution of feature distributions across all 40 classes. Bottom: Detailed depiction of the evolution of feature distributions for select typical classes.}
    \label{fig: feature_vis}
    \vspace{-0pt}
\end{figure*}

\noindent\textbf{Text to 3D shape retrieval.}
Figure \ref{fig: app-TtoP} shows the retrieval results on the Objaverse dataset for text to 3D shape retrieval. 
The retrieved 3D shapes exhibit a high degree of congruence with the given textual descriptions, effectively capturing both the general structure and specific details. For example, the description "wooden four-tier dresser" yields 3D shapes that accurately reflect the specified material and tier structure. Similarly, the "red mushroom with spots" retrieval demonstrates precise adherence to both shape and color details. The retrieval of "table with books and fruit on it" shows MM-Mixing's capability to capture complex arrangements and specific object placements. These text-to-3D shape examples demonstrate that MM-Mixing significantly enhances retrieval accuracy, providing robust and detailed matches that affirm its efficacy in multimodal retrieval tasks.

\subsection{Point Cloud Feature Distribution}
Figure \ref{fig: feature_vis} illustrates the evolution of high-level feature distributions of the Point-BERT pre-trained on the Ensembled dataset during the training process via t-SNE. In the early stage feature distribution, the feature space is highly scattered with overlapping clusters, indicating that 3D backbone has not yet learned to effectively discriminate between different classes. As the 3D backbone starts to learn more features based on mixing alignment, the transitional feature distribution shows a notable improvement, with clusters becoming more distinct. However, there still remains some inter-class overlap.

In the final feature distribution, the clusters are well-separated and compact, reflecting a highly discriminative feature space. 3D backbone has successfully learned to distinguish between different classes with a high degree of accuracy. The representative clusters at the bottom of each visualization further emphasize this progression, showing a clear transition from mixed and overlapping clusters in the early stages to well-defined and isolated clusters in the final stage.
These visualizations highlight the effectiveness of the MM-Mixing pre-training process, demonstrating a clear trajectory of improvement in feature discrimination, culminating in a robust and well-defined feature space.

\subsection{Limitations Discussion}
While our proposed MM-Mixing method combines input-level and feature-level mixing alignment to balance cross-modal consistency and realistic data variation, there are several limitations to consider.  

On the one hand, dual-level mixing, despite its benefits in generating realistic variations, demands significant computational resources, which might not be feasible for all applications, especially those with limited hardware capabilities. On the other hand, single-feature-level mixing, while computationally efficient, may introduce abstract changes that are less intuitive and might not always capture the full complexity of the raw data. Secondly, our approach assumes the availability of sufficient and diverse training data, which might not be the case in every scenario. Additionally, as faced by many deep learning works, the pre-training performance is somewhat limited by the setting of hyperparameters, and finding the best value is challenging. Lastly, the integration of multiple datasets, as proposed in OpenShape, can introduce inconsistencies and require careful preprocessing to ensure data quality and compatibility. 

These limitations highlight areas for further research and development to enhance the robustness and applicability of our method.

\subsection{Potential positive societal impacts and negative societal impacts}
The advancements in triplet generation for point clouds and the integration of multimodal learning frameworks hold significant positive societal impacts. Enhanced 3D data alignment with other modalities can improve various applications, including autonomous driving, medical imaging, and virtual reality. For instance, better 3D shape descriptions can lead to more accurate medical diagnoses and advanced treatment planning. In the realm of education, these technologies can facilitate more immersive and interactive learning experiences. 

\end{document}